\documentclass[sigconf]{acmart}

\usepackage{graphicx}
\usepackage{subfig}
\usepackage{placeins}
\usepackage{enumitem}
\usepackage{url}
\usepackage{multirow}

\usepackage{flushend}
\usepackage{balance}
\emergencystretch=\maxdimen
\AtBeginDocument{%
  \providecommand\BibTeX{{%
    \normalfont B\kern-0.5em{\scshape i\kern-0.25em b}\kern-0.8em\TeX}}}

\copyrightyear{2023}
\acmYear{2023}
\setcopyright{acmlicensed}
\acmConference[HRI '23]{Proceedings of the 2023 ACM/IEEE International Conference on Human-Robot Interaction}{March 13--16, 2023}{Stockholm, Sweden}
\acmBooktitle{Proceedings of the 2023 ACM/IEEE International Conference on Human-Robot Interaction (HRI '23), March 13--16, 2023, Stockholm, Sweden}
\acmPrice{15.00}
\acmDOI{10.1145/3568162.3576998}
\acmISBN{978-1-4503-9964-7/23/03}

\begin{document}

\title[Crafting with a Robot Assistant]{Crafting with a Robot Assistant: Use Social Cues to Inform Adaptive Handovers in Human-Robot Collaboration}

% \author{Anonymous Authors}

\author{Leimin Tian}
\email{leimin.tian@monash.edu}
\orcid{0000-0001-8559-5610}
\affiliation{%
  \institution{Monash University}
  \city{Clayton}
  \state{VIC}
  \country{Australia}
}

\author{Kerry He}
\email{kerry.he@monash.edu}
\orcid{0000-0003-4052-969X}
\affiliation{%
  \institution{Monash University}
  \city{Clayton}
  \state{VIC}
  \country{Australia}
}
% \authornote{Equal contribution}

\author{Shiyu Xu}
\email{sxuu0041@student.monash.edu}
\orcid{0000-0001-9611-8668}
\affiliation{%
  \institution{Monash University}
  \city{Clayton}
  \state{VIC}
  \country{Australia}
}

\author{Akansel Cosgun}
\email{akan.cosgun@deakin.edu.au}
\orcid{0000-0003-4203-6477}
\affiliation{%
  \institution{Deakin University}
  \city{Burwood}
  \state{VIC}
  \country{Australia}
}

\author{Dana Kuli\'{c}}
\email{dana.kulic@monash.edu}
\orcid{0000-0002-4169-2141}
\affiliation{%
  \institution{Monash University}
  \city{Clayton}
  \state{VIC}
  \country{Australia}
}

% \renewcommand{\shortauthors}{Anon et al.}
\renewcommand{\shortauthors}{Leimin Tian, Kerry He, Shiyu Xu, Akansel Cosgun, \& Dana Kuli\'{c}}

\begin{abstract}
We study human-robot handovers in a naturalistic collaboration scenario, where a mobile manipulator robot assists a person during a crafting session by providing and retrieving objects used for wooden piece assembly (functional activities) and painting (creative activities). We collect quantitative and qualitative data from 20 participants in a Wizard-of-Oz study, generating the \textbf{F}unctional \textbf{A}nd \textbf{C}reative \textbf{T}asks Human-Robot Collaboration dataset (the FACT HRC dataset), available to the research community. This work illustrates how social cues and task context inform the temporal-spatial coordination in human-robot handovers, and how human-robot collaboration is shaped by and in turn influences people's functional and creative activities. 
% Supervised learning experiments on the dataset show that the operator's controls can be predicted from observations of the participants' behaviours, emotions, and task progress, providing a first step towards autonomous and socially adaptive human-robot handover.
\end{abstract}




%   \includegraphics[width=\textwidth]{fig/BirdhouseEnsembleCompressed.pdf}
%   \caption{An ensemble of the 20 birdhouses created by our participants.}
%   \Description{The 20 birdhouses created by our participants have various visual designs with different colours and patterns used.}

%   \centering
%   \includegraphics[width=0.7\textwidth]
%   {fig/ExperimentLayoutWithPeopleAnnoZoom.png}
%   % \vspace{-1em}
%   \caption{Layout of the experimental set-up.}
%   \Description{The participant and the experimenter sit at each end of the room. The participant has a desk and a bin at the work space, while the experimenter has a desk and a box at the storage space. The operator sits behind the participant in a neighbouring room and controls the robot via a teleoperation interface. The Fetch robot moves between the work space and the storage space to bring the objects needed for crafting or for return. An OAK-D video camera is set up to the front right of the participant for additional data capture.}
%   \label{Space_layout}
%   % \vspace{-1em}

\maketitle

\section{Introduction}\label{sec:intro}
\begin{figure}[tb]
  \centering
  \includegraphics[width=\linewidth]
  {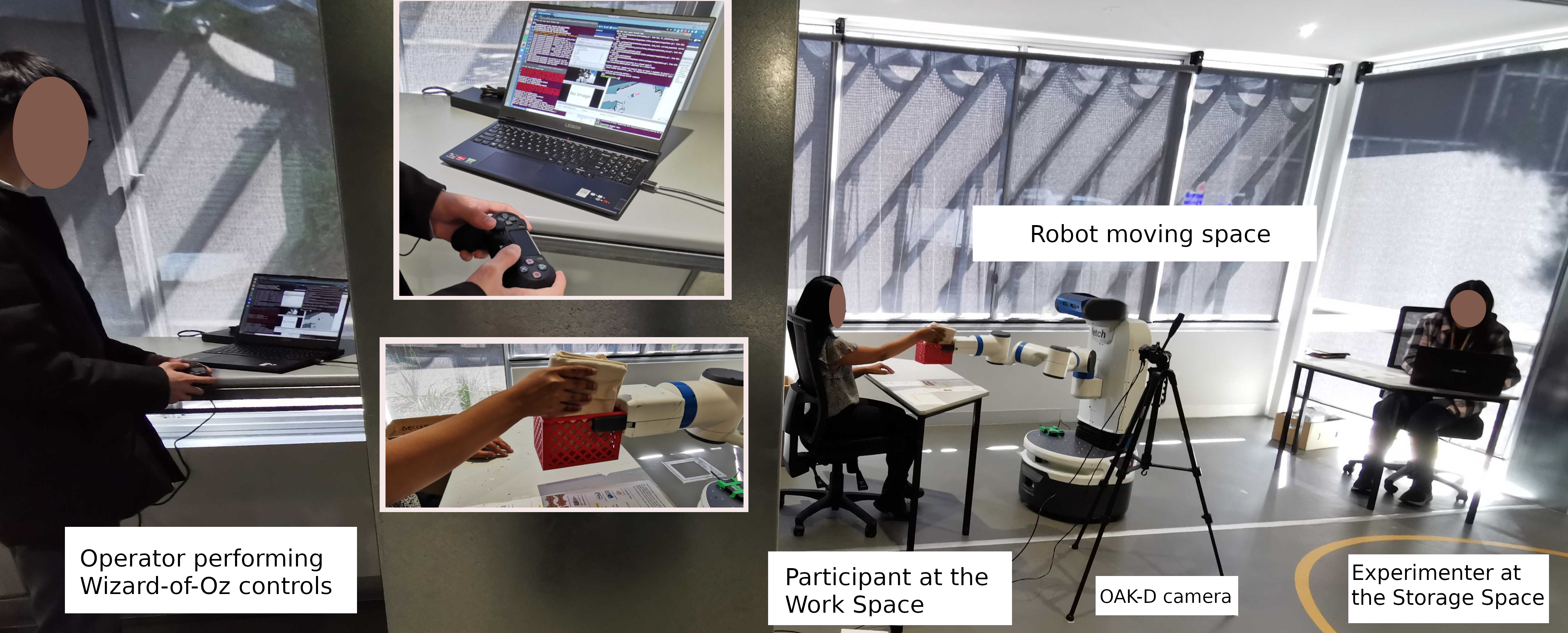}
  % \vspace{-1em}
  \caption{Experimental space layout: the participant and the experimenter sat at the work space and storage space, respectively. The Fetch robot moves in between to pass objects in a basket. The operator sat in a neighbouring room.}
  \Description{The participant and the experimenter sit at each end of the room. The participant has a desk and a bin at the work space, while the experimenter has a desk and a box at the storage space. The operator sits behind the participant in a neighbouring room and controls the robot via a teleoperation interface. The Fetch robot moves between the work space and the storage space to bring the objects needed for crafting or for return. An OAK-D video camera is set up to the front right of the participant for additional data capture.}
  \label{Space_layout}
  % \vspace{-1em}
\end{figure}

Human-Robot Collaboration (HRC) is an active research field that investigates how robots can work together with people as a synergistic team to achieve beneficial outcomes, such as in education~\cite{rosenberg2019human}, healthcare~\cite{pineau2003towards}, or manufacturing~\cite{matheson2019human}. A common HRC task is human-robot handover, which is the physical exchange of an object between a person and a robot at a mutually agreed location and time~\cite{ortenzi2021object}. A successful handover requires spatial and temporal coordination between the person and the robot. Further, how the physical exchange unfolds is influenced by environmental, personal, and task constraints~\cite{parastegari2017modeling}. Current efforts in developing better handovers in HRC have focused on creating models that can identify human intentions and adapt the robot's motion planning and execution accordingly. However, as identified by~\citet{ortenzi2021object}, existing studies investigated handovers in isolation without incorporating them in a naturalistic context with pre- and/or post-handover HRC tasks. Further, handover evaluation has focused on the performance and technical aspects, e.g., handover time and success rate, rather than the social interaction. This limits the understanding of human perceptions and behaviours during handovers, as well as the development of an effective and socially appropriate HRC. 

Our goal is to understand how people perceive a robot and exchange objects with it during task-focused HRC. To build towards autonomous robot capabilities, we conduct a Wizard-of-Oz (WoZ) study to investigate how human operators achieve adaptive, efficient, and socially appropriate HRC. We design a crafting task in which a person assembles and paints a wooden birdhouse with the assistance of a mobile manipulator robot, as shown in Figure~\ref{Space_layout}. The person and the robot exchange a set of task-relevant objects with diverse usage purposes (see Figure~\ref{exp_overview}). We collect quantitative and qualitative data to understand human experience and behaviour during HRC, as well as effective and adaptive collaboration strategies. Recordings of the HRC sessions and the teleoperation framework we developed are made publicly available.
% \footnote{The FACT HRC dataset: \url{shorturl.at/ls347}}
% Temporary anon G-Drive logins: 
% anonrobopsychologist@gmail.com
% Dr.SusanCalvin_the_1st

\section{Background}
HRC relies on an effective and appropriate social communication between humans and robots~\cite{unhelkar2020decision, tian2021taxonomy}. Expression and perception of social cues is key to successful HRC, for instance, using hesitation gestures to negotiate access to shared resources~\cite{moon2021design}. Similarly, previous research found that people use body stances and arm gestures to communicate their intent to when and where a handover happens~\cite{cakmak2011using}, both as the giver~\cite{pan2017automated} and as the receiver~\cite{grigore2013joint}. The quality of human-robot handover can be assessed on various dimensions using both subjective and objective metrics. For example, \citet{murphy2013survey} evaluated handover quality as success rate, human/robot idle time, task completion time, and cost functions relating to movement trajectories, while \citet{hoffman2019evaluating} measured human-robot cooperation and contribution using subjective questionnaires.

Previous studies typically investigate handovers in isolation where objects are exchanged without a pre- and/or post-handover task context, for example, to develop better models for predicting human motions~\cite{wu2019line, chan2020affordance}, to predict the appropriate location of object transfers~\cite{nemlekar2019object, maeda2017probabilistic}, to create more natural and fluent robot motions~\cite{bijl2017system, chan2021experimental}, or to develop control models adaptive to contextual factors, such as human activities, preferences, and object semantics~\cite{kupcsik2018learning, karampatziakis2020empirical, feng2020high}. In these studies, it is unclear how a person's perception and behaviours during handovers may change when they are integrated as part of an HRC task. 

An example study that investigated handovers as embedded in a collaborative task context is \citet{Huang2015adaptive}. The authors studied timing coordination in human-human handovers when they collaborated at a task of unloading a dish rack. Based on observations of 8 pairs of participants, \citet{Huang2015adaptive} developed and evaluated a rule-based model for a robotic arm mounted on a fixed table, which adapted its movement speed and pause duration in between robot-to-human handovers to replicate human decisions during the dish unloading task. This work provided valuable insights on temporal coordination in handovers. However, further research is required to understand human-robot handovers as a social interaction for both temporal and spatial coordination, as well as for diverse types and goals of the object transfer.

\section{Methodology}\label{sec:method}
% We conducted WoZ experiments and data collection to investigate handovers within a naturalistic HRC context. 
Here we provide an overview to our study design, including human research ethics, experimental protocols, custom robot teleoperation framework, and collection of measurements.

\subsection{Human Ethics and Data Collection}
Our study protocols were reviewed and approved by the Monash University human research ethics committee (No.31927). We recruited 20 participants from the university's staff and student population (10 females and 10 males, age $27.00\pm5.17$): fifteen were from the faculty of engineering, four from the faculty of art (P7, P8, P10, P15), and one (P13) from the faculty of business and economics. All participants provided informed consent for taking part in the study and having their data recorded as part of a public dataset for research purposes. As a WoZ study, the participants were not told explicitly before the experiment that the robot was teleoperated. Instead, the experimenter explained the purpose and details of the WoZ approach in the debriefing afterwards.

\subsection{Experiment Protocol}\label{subsec:studyI}
\begin{figure*}[tb]
  \centering
  \includegraphics[width=\textwidth]{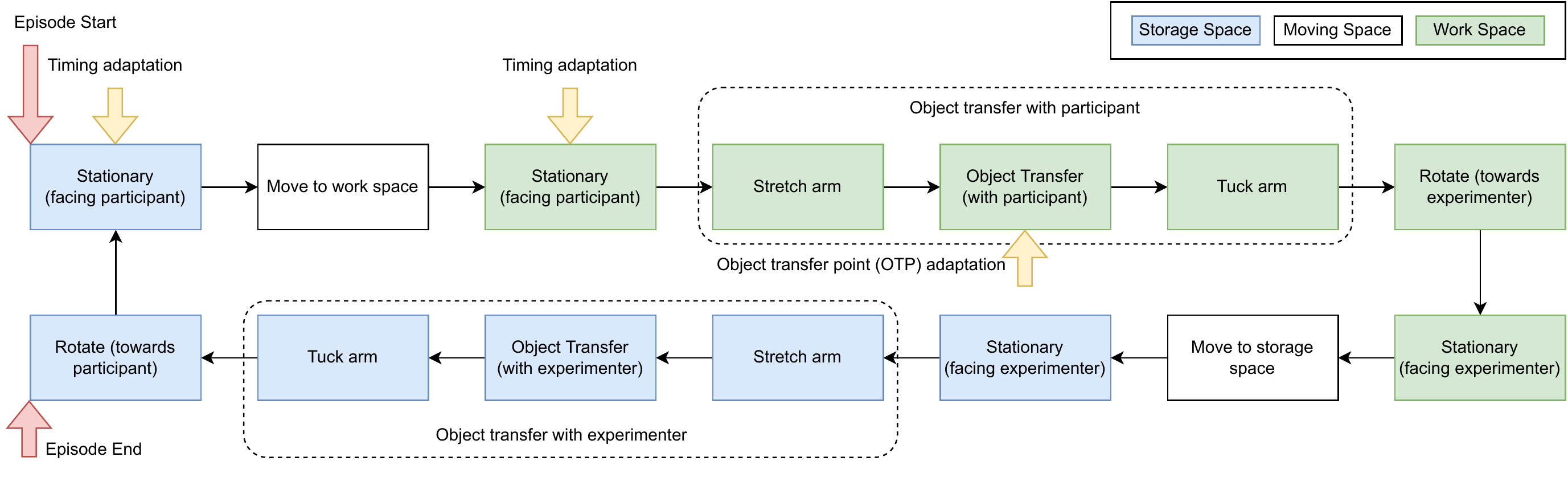}
  \caption{An episode of handover: the operator performs temporal adaptation when waiting to move the robot to the work space and when waiting to initiate object transfers with participant. Spatial adaptation also happens during object transfers.}
  \Description{An episode includes object transfer with participant and with experimenter. It begins with the robot moving to work space to perform object transfer with with participant and ends with the robot finishing object transfer with experimenter and becoming ready for the next handover episode. Spatial adaptation of handovers happens during object transfer with participants. Temporal adaptation of handovers happens when the robot is waiting to move to the work space and when the robot is at the work space waiting to initiate object transfer with participant.}
  \label{episode_overview}
\end{figure*}

In the collaborative task, a person assembles and paints a birdhouse/feeder with the assistance of a Fetch mobile manipulator. The participants kept their creation as an appreciation for their time, which is expected to increase their engagement in the task. A session involves three human roles, as shown in Figure~\ref{Space_layout}: a \textit{Participant} sitting at a work space at one side of the room with a bin and a desk too small to fit all the required objects at once; an \textit{Experimenter} sitting at a storage space at the other side of the room with all objects required during the session stored in a box; an \textit{Operator} sitting behind the participant hidden from their view who controls the robot. The robot moves in between the work space and storage space and performs object handovers on both sides. The detailed state machine during each handover episode is shown in Figure~\ref{episode_overview}. The robot (and operator) observes the participant using its onboard head-mounted camera, as well as an RGB-D camera (OAK-D) installed next to the work desk to the front right of the participant.
An instruction sheet listing all the objects used in a session is displayed on the top right corner of the work desk for the participants to refer to at any time.
% , which lists all the relevant objects in order (with photos), what they are for, whether they need to be returned or not, an example of a finished unpainted birdhouse, and a QR code linking to the online questionnaire.
Note that because we are interested in identifying how a robot can best provide assistance in terms of handover timing and object transfer point (OTP), we simplified the handover problem: The robot holds a basket in which objects are transferred, rather than using its gripper to grasp the various objects directly.
% (see FACT-support in Section~\ref{subsec:FACT})

%\footnote{\url{https://fetchrobotics.com/fetch-mobile-manipulator/}} (Figure~\ref{Fetch_robot})
%\footnote{Laser-cutting source file see FACT-support (design adapted from \url{https://3axis.co/laser-cut-decorative-wooden-bird-feeder-dxf-file/lopq8r57/})}
% (Figure~\ref{fig_CAD})
% The birdhouse consists of 7 wooden pieces that are laser-cut from 6mm birch sheets. The assembled birdhouse measures approx. 15cm (W) x 15cm (D) x 25cm (H). The Fetch robot used in the experiments has a mobile base, an extendable torso, a 7-DOF arm with a gripper that has a maximum grip force of 245N, and a head with cameras.

\begin{figure}[tb]
  \centering
  \includegraphics[width=\linewidth]{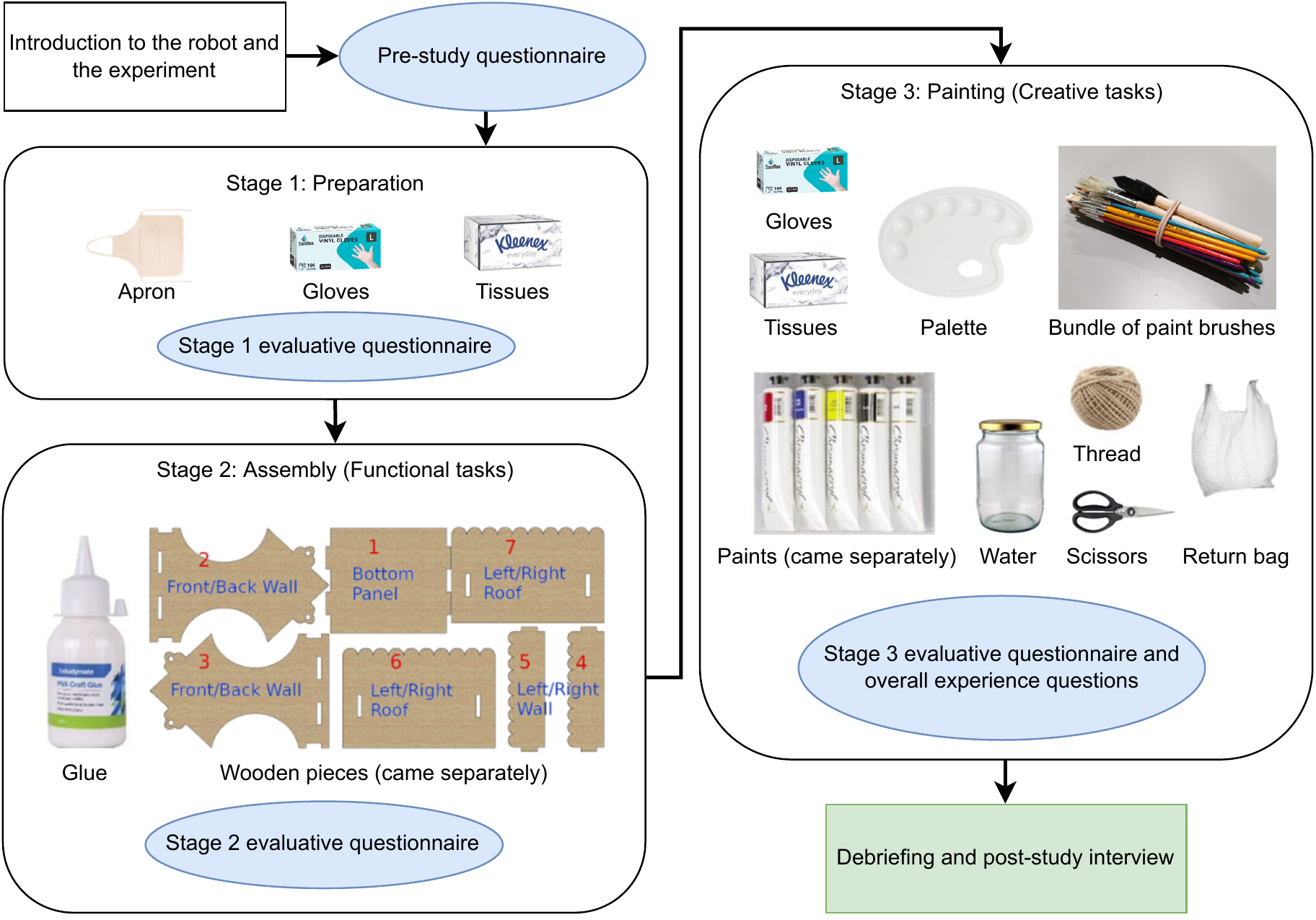}
  % \vspace{-1em}
  \caption{Session overview: the one-hour session includes three stages with various objects used during each stage. Participants filled in an online questionnaire during the session. A debriefing and interview is conducted at the end.}
  \Description{The experiment begins with an introduction to the robot and the session, followed by the preparation, assembly, and painting stages with various object handovers during each stage. Participants filled in an online questionnaires after the introduction and each stages. A debriefing and interview is conducted at the end. The total duration of a session is approximately one hour.}
  \label{exp_overview}
  % \vspace{-1em}
\end{figure}

%an object transfer without pause takes 12s to complete, a full episode without pause takes 42s, the robot adjusting its arm configuration to shift between left and right handover and then performing the object transfer takes 30s. 
The robot-assisted crafting task consists of three stages: Stage 1 (Preparation), Stage 2 (Assembly), and Stage 3 (Painting). The robot and the participant engaged in handovers during each stage, in which the participant can either be the receiver when the robot brings them an object from the storage space, or be the giver when they need the robot to return an object to the storage space. Figure~\ref{exp_overview} provides an overview of the experiment session, which has a total duration of approximately one hour. Stage 1 is aimed at preparing for the collaborative task and an implicit mutual learning between the participant and the robot (operator). Stage 2 is aimed at understanding handovers and HRC in a \textit{functional} task context. Stage 3 is aimed at understanding handovers and HRC in a \textit{creative} task context.
The functional and creative activities have different pre- and post-handover task context, which facilitates our goal of investigating contextualised handovers. In functional activities, participants are expected to focus on assembling the birdhouse quickly and ensuring its structure is sound. In creative activities, participants are expected to have different mental (i.e., design birdhouse appearance to their liking) and behavioural (i.e., realising this personal design by painting) context, which influences how and when the handovers should happen, and participants’ perception and experience during the tasks.

An operator controlled the robot via a teleoperation interface detailed in Section~\ref{sec:teleopGUI}. The operator sat in a neighbouring room (see Figure~\ref{Space_layout}) and took their position after the participant and the experimenter were seated. As a WoZ study, the participants were not informed of the operator's presence before or during the experiment. This allows us to understand what social cues people express during HRC and how these social cues can be used to achieve adaptive handovers applicable to autonomous robots. 

\subsection{Teleoperation Framework}\label{sec:teleopGUI}
We developed a custom teleoperation framework using the Robot Operating System (ROS).\footnote{The teleoperation framework: \url{https://github.com/tianleimin/fetch-teleop}} One expert Fetch operator was trained on using the teleoperation framework, and controlled the robot during all experimental sessions. The operator's role is to identify the appropriate timing for initiating and completing each handover episode, to identify the appropriate OTP, and to evaluate the quality of a handover episode. Further, after each session, the operator participated in the interviews to describe their strategies and to incorporate the participants' feedback. 
% The operator kept a copy of the instruction sheet and a cheat-sheet of the teleoperation framework's controls during the experiments for information.

\begin{figure}[tb]
  \centering
  \includegraphics[trim={0 2.5cm 7cm 0}, clip, width=0.9\linewidth]{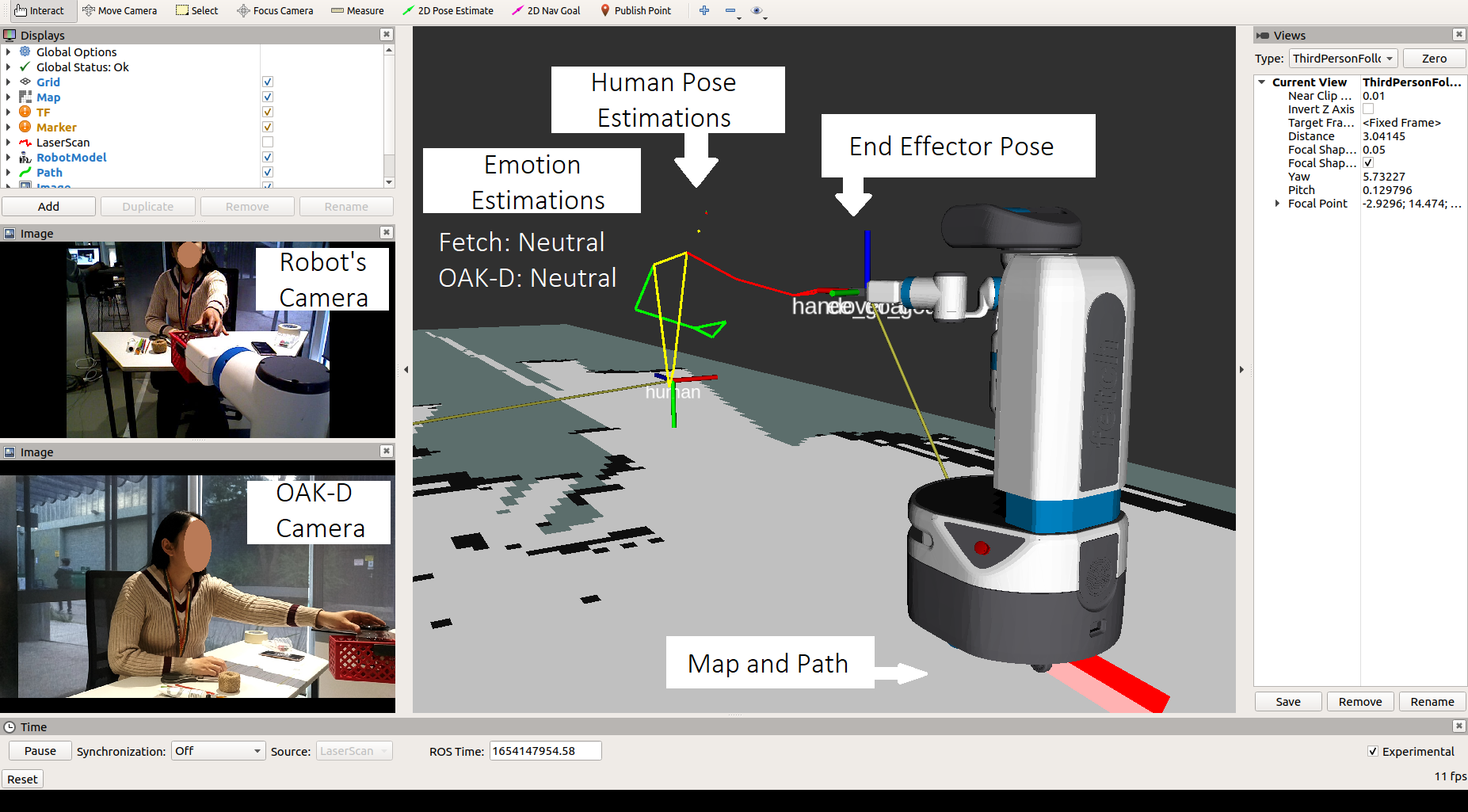}
  % \vspace{-1em}
  \caption{Visualisation of the teleoperation framework: the operator can see live feeds of the robot's onboard camera, the additional RGB-D camera, a 3D map view with the robot's pose as well as estimated participant's poses and emotions.}
  \Description{The operator can see a live feed of the robot's onboard camera and the additional OAK-D RGB-D camera, a 3D map view with the participant's estimated poses and the robot's end effector pose marked, the robot's location along its path in between the work space and storage space, and text outputs of the emotions estimated from the robot's onboard camera and the OAK-D camera.}
  \label{Fetch_teleop_GUI}
  % \vspace{-1em}
\end{figure}

The operator controlled the robot using a hand-held controller and a computer interface. As shown in Figure~\ref{Fetch_teleop_GUI}, the operator has access to a live feed of the OAK-D camera and the robot's onboard camera, as well as estimates of the participants' upper body and facial keypoints from the OAK-D camera feed using Blazepose~\cite{bazarevsky2020blazepose}, and categorical emotion estimates from both camera feeds using EmoNet~\cite{toisoul2021estimation}. The participant's pose, the robot's pose, and the robot's end effector's (EE) pose are shown in a 3D map view, which plots the path that the robot follows to travel between the work space and the storage space. 
% This visualisation interface was implemented using RViz. 
The operator can drive the robot along a predefined path and rotate the robot's base. They can also make fine adjustments to the positions of the path's endpoints during or in between episodes. To perform handovers, the operator can extend the EE towards an OTP (arm stretching), or retract it to a default tucked position (arm tucking). The handover motion sequence is based on \citet{he2022go}. We specified three default OTPs: to the left of the robot (right-hand-side of a participant), to the right of the robot (left-hand-side of a participant), and in the centre. The position and orientation of these OTPs can be fine-tuned by the operator. Changes in the path endpoints and OTPs can be saved or reset. After each episode, the operator uses the controller to evaluate the quality and specify the nature of the participant-robot handover (human-to-robot, robot-to-human, or bidirectional). During arm stretching, the robot automatically tilts its head downwards to point at the EE and display joint attention~\cite{moon2014meet, zheng2015impacts}. During arm tucking, the robot restores its head to neutral position. 

% \begin{figure}[tb]
%   \centering
%   \includegraphics[width=\linewidth]{fig/TeleopDefault}
%   \caption{Controls of the teleoperation framework: Default mode}
%   \Description{The operator can give quick evaluation of the handover, specify whether it is a human-to-robot or robot-to-human handover, reset the handover configuration to default, or save an updated handover configuration.}
%   \label{Fetch_teleop_default}
% \end{figure}

% \begin{figure}[tb]
%   \centering
%   \includegraphics[width=\linewidth]{fig/TeleopBase}
%   \caption{Controls of the teleoperation framework: Base mode}
%   \Description{The operator can change the location and orientation of the robot's base for handover. The changes will be applied to all the following handovers unless reset to default.}
%   \label{Fetch_teleop_base}
% \end{figure}

% \begin{figure}[tb]
%   \centering
%   \includegraphics[width=\linewidth]{fig/TeleopArm}
%   \caption{Controls of the teleoperation framework: Arm mode}
%   \Description{The operator can change the location and orientation of the robot arm's end effector (EE) for handover. There are three default EE locations for left-handed, right-handed, and straight-on handovers. The changes will be applied to all the following handovers unless reset to default.}
%   \label{Fetch_teleop_arm}
% \end{figure}

\subsection{Measurements}\label{subsec:meas}
Data from the teleoperation framework is recorded continuously using ROS bags, which includes synchronised operator control signals, video recordings, robot status, estimates of facial and upper body keypoints from the OAK-D camera feed, and the emotion estimates from both camera feeds. Further, we collected participants' responses to a questionnaire. This includes pre-study questions on participants' demographic backgrounds and robot self-efficacy~\cite{robinson2020robot}. After each of the three stages, the participants evaluated their experience during that stage with 5-point Likert scale measurements of fluency, trust, and working alliance between the themselves and the robot~\cite{ortenzi2021object}, their perception of the robot as the GodSpeed questionnaire~\cite{bartneck2009measurement}, their overall enjoyment and satisfaction, and a text comment box. After all stages are completed and rated, before the debriefing and interview, participants also chose whether they thought the robot was teleoperated or fully autonomous, whether they thought the robot was adaptive, and gave any additional comments in a text box. 

In addition to the quantitative data captured by the questionnaire and the teleoperation framework recording, we collected qualitative observation notes and post-session interviews. The experimenter took observation notes during the session on participants' behaviours, operator's controls and adaptations, and events of interest. In interviews participants were asked to describe their impression of the robot and the HRC experience, their guesses as to how the robot functioned, their reasoning behind certain events of interest during the session, their design and creative processes, and future improvements that the robot can incorporate to become a better assistant. The operator also joined the interview to describe their observational and adaptive strategies, and to answer any questions from participants.

% \section{Socially Adaptive Handover}\label{sec:model}
% Here we describe the teleoperation framework developed for WoZ data collection, as well as the models used in the preliminary machine learning experiments with the collected data.

\section{The FACT HRC dataset}\label{subsec:FACT}
We collected a dataset of WoZ controlled human-robot handovers and collaboration in \textbf{F}unctional \textbf{A}nd \textbf{C}reative \textbf{T}asks (FACT). The FACT HRC dataset includes three segments:
\begin{enumerate}
    \item \textit{FACT-raw}\footnote{FACT-raw: \url{https://doi.org/10.26180/21671789.v1}} (requires a signed EULA to access) contains identifiable ROS bag recordings described in Section~\ref{subsec:meas}.
    \item \textit{FACT-processed}\footnote{FACT-processed and FACT-support: \url{https://doi.org/10.26180/21671768.v1}} contains non-identifiable csv data extracted from the synchronised raw data at intervals of 0.1s (in total approximately 600K data instances), including the robot's status, the operator's controls, the facial and upper body keypoints estimated from the OAK-D camera feed, and the emotion estimates from both camera feeds. In addition, the de-identified questionnaire responses are provided. 
    \item \textit{FACT-support} contains supporting materials, including the instruction sheet, operator's cheat-sheet for teleoperation framework controls, the questionnaire, CAD file for laser cutting the birdhouse pieces, implementation of the teleoperation framework and data processing scripts. 
\end{enumerate}
% FACT-WoZ (Study I) and FACT-Auto (Study II\&III)
% As discussed in Section~\ref{sec:method}, we collected data from 20 participants and the processed csv data at a granularity of 0.1s, resulting in a total of approx. 600K data instances. 
The FACT HRC dataset includes 565 handover episodes, namely:
\begin{itemize}
    \item 52 Human-to-robot handovers (9.2\%): the person giving an object to the robot by placing it in the empty basket that the robot holds in its gripper.
    \item 320 Robot-to-human handovers (56.6\%): the person receiving an object from the robot by taking it from the basket.
    \item 193 Bidirectional handovers (34.2\%): object exchange between the person and the robot, either simultaneously or in two consecutive give and receive motions, during which the robot's arm can remain stretched out or be tucked to wait between the give and receive motions.
\end{itemize}
% , e.g., the person takes the blue paint with one hand and returns the red paint with the other hand simultaneously, or the person takes the glove box out of the basket and returns it shortly after.

% \section{Analysis Discussion}\label{sec:result}
% Here we analyse the operator's adaptation strategies and the participants' perceptions and behaviours from our dataset.

\section{Operator's Adaptation Strategies}\label{subsec:operator}
The operator adapted handovers based on participants' intent and preference inferred from gaze and gestures, safety considerations of the handover, and its efficient integration with the task context. 

%Further, the operator's evaluation of the handover quality did not always align with the participants' perception, with the participants often ignoring minor handover planning and execution mistakes as they were focused on the overall collaborative task beyond object transfers. These strategies can guide the design of a fully autonomous handover model to achieve better human-robot interaction and collaboration.

\subsection{Observation and Expression of Social Cues}\label{subsubsec:operator_observation}
% The operator monitored the participants and the interaction flow to decide on the appropriate controls and to evaluate the handover episodes. 
% As shown in Figure~\ref{Fetch_teleop_GUI}, the operator monitored the interaction via the two camera feeds, the pose visualisations on the map, and print-outs of the facial-expression-based emotion recognition model. 
%When the robot is at the work space facing the participant, the operator focused on the robot's onboard camera feed, while when the robot is rotating or facing the experimenter, the operator relied on the OAK-D camera's feed as the participant is out of view of the onboard camera. 
The operator reported monitoring the person's gaze (e.g., eye contact with the robot), hand gestures (e.g., waving), and upper body movements (e.g., leaning forward), which aligns with existing research \cite{cakmak2011using, ortenzi2021object}. 
The operator did not rely on the estimated emotion print-outs, as the automatic model is less accurate than a human's performance. %The automatic model frequently mistook people's facial expressions as negative emotions (e.g., sadness) when they were focused on the task or had a high cognitive effort. Moreover, it recognised smiles that people displayed as happy when the robot exhibited unnatural or unexpected motions. 
However, emotion estimations may benefit an autonomous model as they can distinguish handover episodes rated as good~vs.~bad by the operator, with more positive or neutral emotions predicted in the episodes rated as good. 
% Namely, for episodes rated as good, 10 participants received majority emotion class prediction (estimated by the robot's onboard camera during robot-participant object transfers) as a positive or neutral category, while for episodes rated as bad, only 4 participants had majority class prediction as positive or neutral.
In addition, the operator monitored pre- and post-handover context, including object layout on the work desk, the space where the person is performing the crafting tasks in, and whether or not one or both of their hands are occupied.
The operator used the position and movement of the robot to communicate its intention to the human collaborator. When the robot is waiting to deliver or receive an object to or from a participant, if the waiting time is estimated to be long, the operator controlled the robot to wait at the storage space to avoid pressuring the person to rush through their tasks, as suggested by some participants in the interviews (Section~\ref{subsubsec_colab}); if the wait is estimated to be short, the operator controlled the robot to wait at the work space so that it could perform the handover as soon as the person is ready. Five participants commented on the robot's head tilting behaviour during object transfers as an indication of the robot's awareness to itself and its collaborator.
% : one perceived it as the robot inspecting where the person's hand was and 4 perceived it as the robot inspecting its own motions.
The operator's observation and control strategies suggest that perception and expression of social cues is key to facilitates HRC.
% e.g., the person is focused on painting, 
% e.g., the person is putting on the apron, 
% (arm tucked and facing the person)
% with P2 perceiving it as the robot inspecting where the person's hand was, and P7, P12, P13, and P15 perceiving it as the robot inspecting its own motions.
% (arm tucked and facing the person)
% As discussed in Section~\ref{sec:teleopGUI}, during object transfers, once the EE reaches the goal location, the robot automatically tilts its head to look at the EE as part of the scripted handover motion sequence to create a sense of shared attention. 

\subsection{Adaptation in Handover Timing}\label{subsubsec:operator_timing}
%The operator's goal for handover timing adaptation is to ensure that the person has a fluent and efficient collaboration with the robot. 
The operator focused on choosing the appropriate moment to move the robot to the work space for initiating the handover episode (i.e., deciding the wait time at the storage space) and to stretch out the robot's arm for initiating the object transfer within a handover episode (i.e., deciding wait time at the work space), as shown in Figure~\ref{episode_overview}. A handover was initiated when the participant displayed a personal ``start'' gesture (Section~\ref{subsec:user_behav}). In addition to these participant-initiated handovers, the operator performed robot-initiated handovers in Stage 2 and 3 by moving the robot to the work space before the participant signalled, when they anticipated that it will take the person shorter to finish their task at hand than the time (9s) the robot needs to move from the storage space to the work space. Robot-initiated handovers were also performed when the item to be delivered is of immediate need to the person while they were continuing their task at hand . 
% The understanding of such initiation cues is established implicitly in Stage 1 (preparation).
% (e.g., consecutive wooden pieces in Stage 2)

% For object transfer and handover completion,  i.e., deciding when to tuck the robot's arm, in robot-to-human or human-to-robot handovers the operator controlled the robot to start tucking its arm once they saw the person has taken/put an object in the basket from the camera feeds, which can happen before or immediately after the EE reached the goal location; in bidirectional handovers, the operator controlled the EE to remain at the goal location to wait until the person has finished the give and receive motions before starting to tuck the robot's arm. However, if the person was to perform actions with the item using the same area of the work desk that the EE is occupying (e.g., putting paint on the palette put in the middle of the work desk while the robot performed a central handover), the arm was tucked in between the give and receive motions of the person to make space for the task. 

The average duration of a handover episode in Stage 1 is 113.6$\pm$42.1s, in Stage 2 is 97.1$\pm$22.8s, in Stage 3 is 108.1$\pm$22.0s, i.e., the episodes are the shortest in the functional activities of assembly (S1~vs.~S2 $p=0.03$, S2~vs.~S3 $p=0.02$, S1~vs.~S3, $p=0.26$). The larger variance in episode duration in Stage 1 is in line with our discussion in Section~\ref{subsec:studyI} on it being intended for implicit mutual learning between the participant and the operator. 
In Stage 1 the average pause at the work space (robot waiting at the participant's side for the right time to initiate object transfer) within the handover episodes is 7.5$\pm$3.8s, and the average pause at the storage space (waiting for the right time to move towards the participant) is 47.9$\pm$24.4s; In Stage 2, the average pause at the work space is 19.0$\pm$15.6s, and at the storage space is 26.3$\pm$14.7s; In Stage 3, the average pause at the work space is 16.2$\pm$13.6s, and at the storage space is 40.1$\pm$27.9s, i.e., the operator controlled the robot to wait longer at the storage space than at the work space ($p\ll0.01$ in S1, $p=0.08$ in S2, $p=0.004$ in S3) to avoid pressuring the participants, in line with our discussion in Section~\ref{subsubsec:operator_observation}.

\subsection{Adaptation in OTP}\label{subsec:OTP}
The operator considered efficiency and safety when deciding on the OTP. For efficiency, they aimed at stopping the EE approximately where the person's dominant or free hand is positioned, or where they signalled the ``start'' gesture. For safety, they factored in object layout on the work desk to avoid collision, especially with the birdhouse that the person is crafting. When the two considerations conflicted the operator prioritised safety. OTP adaptations can happen live in response to a participant's behaviours or between object transfers based on past observations. 
% When adapting between episodes, the operator used the storage side (robot-experimenter handovers) as a preview.

While the operator can move the robot's base closer or further away from the person when it is at the work space, they chose not to use this function. Instead, OTP was adapted by changing the EE goal. As described in Section~\ref{sec:teleopGUI}, there are three default EE goals (left, centre, right). The operator can make changes to these default OTPs. However, such fine adjustments were not performed as one of the three defaults was considered close enough to the operator's envisioned OTP. 
Further, the operator stopped the arm stretching motion prematurely before the default EE goal was reached (i.e., the OTP is further away from the person than the default) if the person reached out their hand(s) before the arm stretching completes.
% Another reason that the operator avoided fine adjustments is that a new arm movement trajectory is automatically generated for the updated EE goal, which can lead to unpredictable or unnatural arm movements that degrade how the robot is perceived by the human collaborator or disrupt the interaction flow. 
% (e.g., moving a short distance above the default goal location)

It takes 12s for the robot to complete an object transfer with the arm ready for one of the three default OTPs or when it is switching between central and left/right OTPs. However, switching between the left and right OTPs takes 30s with bigger motions. To avoid interrupting the collaboration flow, the operator did not perform such left-right switches during robot-participant handovers. Instead, this was done during robot-experimenter handovers or while the robot was waiting at the storage space. In comparison, switching between central and left/right OTPs was performed more frequently in robot-participant handovers to quickly adapt to the person's preferences. Out of all 61 OTP switches, 49 were left-middle switches (80.3\%), 7 were right-middle switches (11.5\%), and 5 were left-right switches (8.2\%). In the 565 handover episodes, 186 (32.9\%) were left handover (right-hand-side of the participant), 367 (65.0\%) were central handover, 12 (2.1\%) were right handover. The preference of left OTP over right OTP aligns with our participant population with only P1 being left-handed. 

\subsection{Operator's~vs.~Participant's Perceptions}\label{subsubsec_OvsP}
Out of all handover episodes, the operator rated 397 (70.3\%) as good, 32 (5.7\%) as bad, and 136 (24.1\%) as neutral.
The operator incorporated the participants' feedback provided in the debriefing and interview after each session in their observational and adaptive strategies in later sessions. Thus, we saw a continual development of the operator's expertise and control strategies:
With P1 and P2, the operator rated the sessions as having the most number of bad handover episodes (5 episodes each). Further, there was one collision incident with P1 where the robot hit the work desk in the first episode due to a setup issue, and one collision incident with P2 in Stage 3 where the robot pushed the assembled birdhouse when stretching its arm to perform central handover to deliver the scissors to P2. P1 and P2 addressed the collision incident during interview, but did not consider the other bad episodes to stand out. With P5, P7, and P12, the operator rated the sessions as having 3 bad episodes each, while P5 considered there were ``one or two miscommunications'', P7 did not notice the robot's mistakes or its changes in OTPs, and P12 commented on ``one weird incident'' with the robot's arm movement appearing unnatural. For remaining sessions with one or two bad episodes as rated by the operator, participants did not address these episodes in the interviews.

% ``one or two miscommunications but otherwise pretty smooth''
% With P3, P4, P9, and P18, the operator rated the sessions as having 2 bad episodes, while the participants did not consider any of the bad episodes to stand out from the rest of the episodes. With P8, P13, P14, P19, and P20, the operator rated the sessions as having 1 bad episode. P13, P14, P19, and P20 did not report any bad handovers. P8 described that she used different hands to signal to the robot which side she preferred the handover to happen, which the operator took notice and understood. However, as the operator prioritised safety (reducing the possibility of colliding with the birdhouse) over user preference when the two conditions conflicted, the handovers did not always align with where P8 signalled. P8 noticed this during the session and guessed that it was due to safety considerations. Thus, she commented that she was ``90\% satisfied that the robot tried to find the best place for doing things''. 

\begin{figure}[tb]
  \centering
  \includegraphics[width=0.9\linewidth]{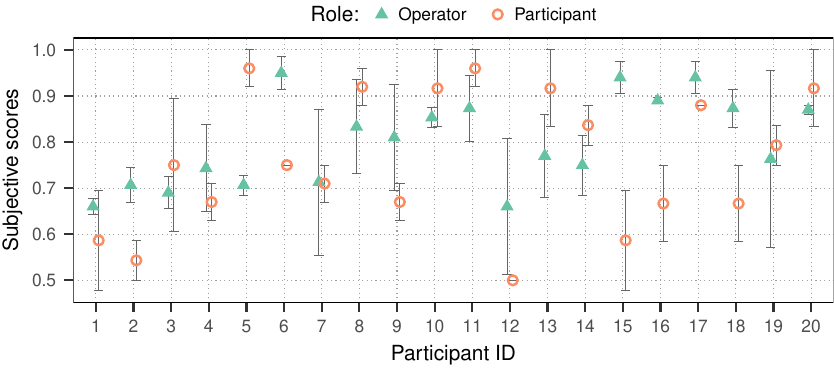}
  % \vspace{-1em}
  \caption{Operator's~vs.~participant's perceptions: the operator had a handover-centred perspective while the participants were influenced by the task context.}
  \Description{The participant's and the operator's ratings of the overall quality of the sessions do not necessarily align, as the operator had a handover-centred perspective, while the participant's perception is more influenced by the HRC task context.}
  \label{fig:OvsP}
  % \vspace{-1em}
\end{figure}

To further understand the differences in operator's and participants' perceptions, for each stage, we calculated an operator's handover quality score $Q_o = 0.5 \times (\frac{N_{good} - N_{bad}}{N_{total}} + 1)$, and a participant's handover quality score as the min-max normalised sum of their ratings on the enjoyable $E$ and satisfactory $S$ level $Q_p = \frac{E + S - 2}{10-2}$ ($Q_o, Q_p \in [0,1]$). Figure~\ref{fig:OvsP} shows the stage-wise mean ($\overline{Q}$) and standard deviation ($\sigma_o, \sigma_p$) of $Q_o$ and $Q_p$ in each session. Note that the participant's ratings are independent from one another, while the operator's ratings of a session can be dependant on their experiences in previous sessions. 
As shown here, the operator and participants' perception of handover quality do not always align. In 9 sessions where $\sigma_o>0.1$ 
(P4, P7-P9, P11-P14, P19), $\overline{\sigma_p}=0.07$, $\overline{\sigma_o}=0.20$. This suggests that compared to the operator who had a handover-centred perspective, the participants' perception may be less influenced by the individual handovers compared to the task context. This is also found in the interviews as discussed in Sections~\ref{subsubsec:robot} and~\ref{subsubsec_colab}.

Comparing $Q_o$ and $Q_p$ in earlier (P1-P10) vs. later (P11-P20) sessions, the average scores in each of the three stages are consistently higher in the later half, although the difference is only statistically significant in Stage 2 of $Q_o$ ($p=0.03$). The increase in the quality scores is bigger for $Q_o$ than for $Q_p$. This indicates that the operator gained expertise in handover adaptation, which improved the participants' experience in later sessions.

% Pearson's correlation coefficient $r$ between the operator's and participant's scores (average of three stages) over all sessions is 0.22 (low).
% Operator's overall score for all sessions = $0.82\pm0.10$. Participant's overall score for all sessions = $0.76\pm0.15$
% $r$ for stage 1 and stage 2 are both 0.12, for stage 3 is -0.05
% The operator attended the interviews and clarified with the participants on (mis)alignment in the operator's intention and the participant's perception, as well as strategies to improve their controls. For sessions in which the operator considered some handover episodes to be bad or included mistakes, the participants did not necessarily notice or remember the bad handovers during the interview. 

\section{User's Perception and Behaviours}\label{subsec:user}
We investigate how people collaborate with and perceive the robot from subjective questionnaires, observation notes, and interview data. Our analysis showed that participants have both positive and negative impressions towards the robot and the HRC experience. Their perceptions are more positive in Stage 2 and 3 compared to Stage 1, as the operator adapted to their preferences and the participants became more focused on the functional and creative activities beyond object handovers. Further, their social cues change within a session and differ between participants. Moreover, the HRC directly influenced people's design and creative processes, even though only human-robot object handovers were performed. 
%Our findings contribute to the understanding of HRC in functional and creative tasks, which will facilitate human-centred design of robot behaviours to support an adaptive, fluent, and socially appropriate HRC, as well as inform human modelling for simulation-based studies.

\subsection{User Behaviours During the HRC}\label{subsec:user_behav}
Interestingly, all participants signalled to the robot's ``face'' (onboard camera), even though they were aware that the OAK-D camera was also used for monitoring their behaviours and was positioned closer to them. P6, who has worked with robots before, was the only participant who attempted signalling to both cameras in the first handover episode and switched to only signalling to the robot's onboard camera in all later episodes. P6 commented that he signalled to the OAK-D camera at the start because it had ``better angle'', but later it felt more natural to signal to the robot's onboard camera. Further, P15 commented that at times he felt it was ``weird'' or ``too formal'' to use big gestures to signal the robot to come, as he considered the robot to be an equal partner working on a task together with him in a casual way, while waving your hand at someone's face can be considered impolite. These highlight the social interaction aspect of human-robot handovers. Although they were told that the robot has no speech recognition or interaction capabilities, five participants still spoke to the robot, such as saying ``thank you'' after handovers. This suggests that speech is a natural interactive modality that can benefit HRC.
% These suggest that the participants treated the robot as a social actor.
% Similarly, P8 expressed that she felt guilty to keep the robot waiting, although she did not feel pressured as if a human was ``breathing down your neck'' when the robot was waiting for her at the work space.

When signalling to the robot, participants used social cues combining one or more modalities, including gaze, movements of hand(s), and body gestures. The most common social cue to signal the initiation of handovers, i.e., the ``start'' gesture, is to gaze up at the robot while leaning forward and showing a single-handed wave, either side-to-side or forward-backward, as used by 17 participants. Several variations of this ``start'' gesture were observed, such as using only gaze without hand or body movements, holding the object to be given to the robot when waving, or using semantic gestures to demonstrate the object needed (e.g., showing gestural imitation of putting on gloves to initiate glove box handover).
% 17 participants (except for P2, P6, P11)
% One variation of this ``start'' gesture is to only gaze up at the robot without any hand or body movements. This was used by P2, as well as in later episodes during the assembly stage (P15, P16, P17) and the painting stage (P3, P5, P15, P16, P17). Other hand movement variations of the ``start'' gesture include having one hand (P6, P14, P16) or both hands (P13) stretched out and palm(s) up, having one hand raised up (P11), a two-handed ``come here'' inward wave (P10), having one hand stretched out holding the object they want to give the robot (P12, P14, P15, P17, P19) or holding and waving the object (P13). In addition, 3 participants (P13, P18, P19) used semantic gestures to demonstrate the object they needed, including pretending to put on an apron, pretending to put on gloves, pretending to take objects out of a box in the preparation stage (P13, P18, P19), pretending to spread glue on surfaces in the assembly stage (P13), and pretending to putting objects into a bag in the painting stage (P13). 
% 12 participants (P3, P5, P6, P10, P12-P19) used more than one ``start'' gesture during the session
Twelve participants used more than one ``start'' gestures during the session, with 11 switching to lower-effort ``start'' gestures in later episodes (except for P13), for instance, from using both gaze and hand movements to gaze-only. During the interview, P1, P3, P10, and P13 discussed that they experimented with different gestures and observed the robot's reactions to decide on which gestures were the most effective. P10 commented that she used bigger gestures at the start of the session because she was more excited. P14 expressed that she changed to smaller gestures later in the session as she became more familiar with how the robot functions and felt that she needed less effort to communicate with it. 
After the robot arrived at the work space, five participants displayed a follow-up ``start'' gesture to signal the robot to stretch out its arm and initiate object transfer by reaching out one hand. After completing object transfers, six participants displayed an ``end'' gesture to signal the robot to tuck its arm and return to the storage space, by showing a single-handed gesture of ``OK'' while nodding or an outward ``go-away'' single-handed wave.
% 5 participants (P7, P11, P15, P18, P19) displayed a follow-up ``start'' gesture
% 6 participants (P8, P10, P13, P14, P16, P18) displayed an ``end'' gesture to signal the robot to tuck its arm and return to the storage space, by showing a single-handed gesture of ``OK'' while nodding (P8), or an outward ``go-away'' single-handed wave (P10, P13, P14, P16, P18).

% Although they were told that the robot has no speech recognition or interaction capabilities, 5 participants (P10, P13, P15, P17, P18) still spoke to the robot. These include saying ``thank you'' after handovers (P13, P15), saying ``there you go Fetch'' after human-to-robot handovers (P18), calling to the robot by saying its name ``Fetch'' in the first handover episode while gesturing (P15), saying ``nope'' during the first handover episode when the robot arrived at the work space with arm tucked and then saying ``yep'' when the arm stretched out to signal preference (P17), and saying ``wow'' during the first handover episode when the robot started moving towards the work space after the person gestured (P10). This suggests that speech is a natural interactive modality that can benefit HRC, which aligns with the interview responses of suggesting speech interaction as a possible improvement, as discussed in Section~\ref{subsubsec:robot}.

During the sessions, participants also adapted to the robot's behaviours. As the robot's movements are relatively slow, to increase efficiency, participants performed bidirectional object exchanges instead of single-directional object handovers when possible. They also filled the waiting time with other activities, namely reading the instruction sheet, inspecting their assembly or painting progresses, and checking their phones. Further, participants configured the layout of their work desk dynamically to keep the area that they preferred for object transfers accessible.

Our analysis showed that participants treated the robot as a social actor during the collaboration and used various social cues, namely gaze, hand movements, and body gestures, to coordinate the timings and locations for object handovers. They also adapted to the robot to achieve more efficient collaboration by changing their behaviours, such as filling in the waiting time with other tasks and performing bidirectional handovers when possible.

\subsection{Perception of the Robot}\label{subsubsec:robot}
% In the pre-study questionnaire, participants rated their robot self-efficacy expectations, with operation efficacy = $13.45\pm2.35$, and application efficacy = $12.95\pm2.37$.\footnote{For robot self-efficacy we report sub-item sums, value range [4,20]. For robot and HRC perception we report sub-item means, value range [1,5].} Participants from the faculty of engineering (15) have operation efficacy = $13.60\pm2.59$ and application efficacy = $12.80\pm2.62$. Participants from the faculties of art and business (5) have operation efficacy = $13.00\pm1.58$ and application efficacy = $13.40\pm1.52$. As shown here, participants from the two populations have similar levels of confidence in their expected efficacy in operating the robot and using it as an application.
Before the HRC session, participants rated their expectations on robot operation and application efficacy. No significant differences were found between participants from the faculty of engineering and participants from the faculties of art and business.

Before debriefing, participants guessed whether the robot was autonomous or teleoperated, and whether they thought it was adaptive. Fifteen participants answered the robot was adaptive during the session, three answered it was not adaptive, and two were unsure. Nine participants thought the robot was fully autonomous, nine thought that it was being teleoperated, and two were unsure. None of the participants identified where the operator was located until they were introduced during the debriefing. The main reason participants gave for considering the robot to be teleoperated or to be unsure about their guesses is the robot's ability to adapt to their behaviours and the task context, which they consider to be too difficult for a fully autonomous robot. This demonstrates that improving a robot's adaptive capabilities is key to achieving natural HRC.
% Out of the two participants who were unsure, P6 has worked with the Fetch robot before while P13 has no previous experience with the robot. In the interview, P6 guessed the robot to be semi-autonomous, because based on his experience and understanding of the robot, its adaptive behaviours would be too difficult to implement with a fully autonomous model. P13 explained in the interview that some of the robot's behaviours appeared to follow a pattern, which led her to believe it was autonomous. However, she also noticed that the robot was able to respond to different gestures that she displayed, which she considered more likely to be teleoperated behaviours. None of the participants identified where the operator was located until they were introduced during the debriefing.

\begin{figure}[tb]
  \centering
  \includegraphics[width=0.9\linewidth]{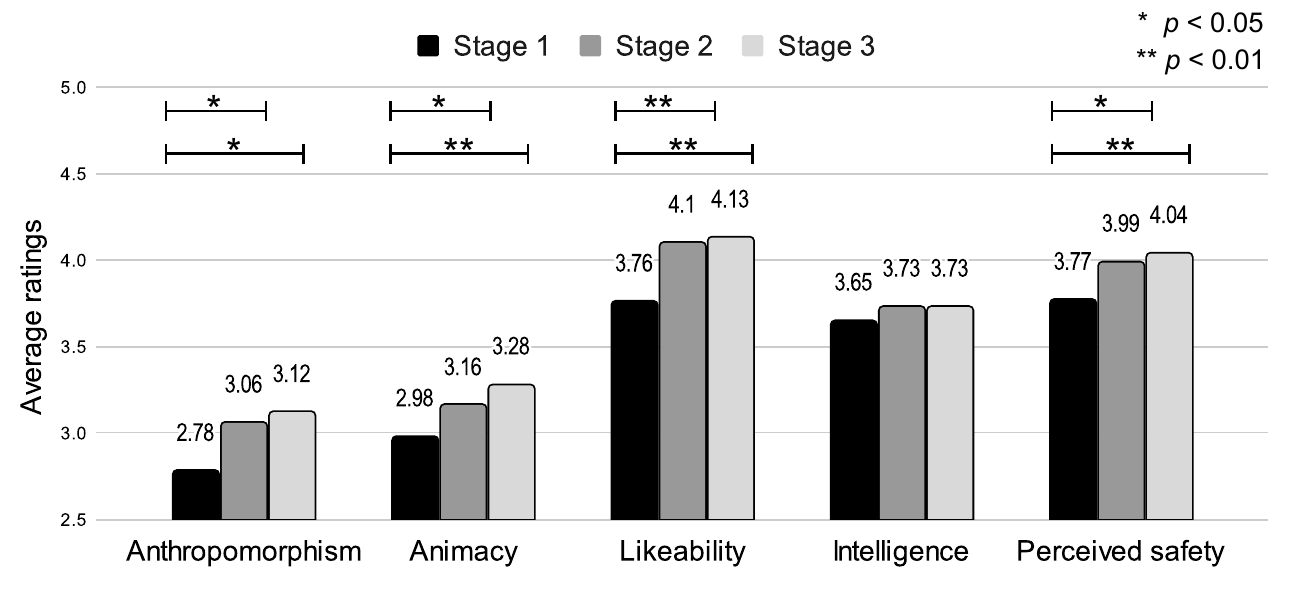}
  % \vspace{-1em}
  \caption{Subjective ratings of robot impression: participants had more positive impressions of the robot in Stage 2 \& 3.}
  \Description{Participants have more positive impressions of the robot in Stage 2 and 3 compared to Stage 1, with Stage 3 more positive than Stage 2, as measured on the dimensions of the robot's perceived anthropomorphism, animacy, likeability, intelligence, and perceived safety.}
  \label{fig:Q_robot}
  % \vspace{-1em}
\end{figure}

In Figure~\ref{fig:Q_robot} and Table~\ref{tab_qualtrics_robot}, we report participants' subjective impressions of the robot in each stage (sub-item means, value range [1,5]). As shown here, participants have significantly more positive impressions of the robot in Stages 2 and 3 compared to Stage 1 (except for Intelligence). One possible cause of the higher ratings in Stages 2 and 3 is that the operator learned an effective adaptation strategy during Stage 1 and performed more fluent handovers in Stages 2 and 3. However, as discussed in Section~\ref{subsubsec_OvsP}, another reason may be the participants being more focused on the functional and creative activities in Stages 2 and 3 rather than object transfers, i.e., the participant's perceptions were influenced more by the collaborative task than by the individual handovers.
% Table~\ref{tab_qualtrics_robot}
% Their impression of the robot in Stage 3 is more positive than Stage 2, although the difference is insignificant.

In the interviews, participants expressed both positive and negative impressions towards the robot. Eight participants commented that the robot was ``cool'', ``impressive'', ``amazing'', ``likeable'', or they were ``fascinated'', ``excited'', or ``satisfied'' interacting with it; Four considered the robot ``responsive'' or ``adaptive'' in its interaction; Three thought the robot was ``useful'', ``effective'', or ``helpful'' for the tasks; Two considered the robot ``reliable'' or ``trustworthy'' in executing actions and performing its tasks. However, eight participants also mentioned slowness and experiences of waiting for the robot. Although this sense of slowness is not necessarily a negative experience for every participant. As discussed by P15, the robot being slow made it feel predictable and non-threatening, which led to him considering the robot ``docile'' and ``cute''. Further, P2 and P18 discussed a sense of uncertainty or confusion in how to interact with or trigger certain behaviours in the robot despite having prior exposure to the robot.

\begin{table}[h]
\begin{center}
\caption{Participants' impressions of the robot and the HRC (mean $\pm$ std). Stage 2 \& 3 are perceived more positively overall.}
\label{tab_qualtrics_robot}
\begin{tabular}{|l|c|c|c|}
\hline
Robot impression & Stage 1 & Stage 2 & Stage 3 \\
\hline
Anthropomorphism & $2.78\pm0.78$ & $3.06\pm0.95$ & $3.12\pm0.96$ \\
\hline
Animacy & $2.98\pm0.70$ & $3.16\pm0.71$ & $3.28\pm0.81$ \\
\hline
Likeability & $3.76\pm0.66$ & $4.10\pm0.74$ & $4.13\pm0.70$ \\
\hline
Intelligence & $3.65\pm0.71$ & $3.73\pm0.66$ & $3.73\pm0.83$ \\
\hline
Perceived safety & $3.77\pm0.68$ & $3.99\pm0.62$ & $4.04\pm0.70$ \\
\hline
\hline
HRC impression & Stage 1 & Stage 2 & Stage 3 \\
\hline
Fluency & $3.62\pm0.74$ & $3.95\pm0.69$ & $4.12\pm0.75$ \\
\hline
Trust & $3.93\pm0.91$ & $4.18\pm0.78$ & $4.20\pm0.85$ \\
\hline
Working Alliance & $3.97\pm0.91$ & $4.09\pm0.76$ & $4.14\pm0.86$ \\
\hline
Enjoyable & $3.70\pm0.86$ & $4.35\pm0.59$ & $4.30\pm0.66$ \\
\hline
Satisfactory & $3.65\pm0.75$ & $4.10\pm0.85$ & $4.10\pm0.85$ \\
\hline
\end{tabular}
\end{center}
% \vspace{-1em}
\end{table}

Participants also discussed future improvements they would want the robot to have in order for it to be a better assistant. Eight participants commented that speech interaction or voice commands in addition to the current gesture-based approach can help increase the interactivity and flexibility of the collaboration. Similarly, three participants suggested that developing a dictionary that associates gestural commands to specific objects can help minimise confusion and increase flexibility. Related to this desire for more flexibility, six participants discussed that it would be good to allow customising the order and set of objects that the robot delivers. Regarding the robot's behaviours, four participants expressed wanting the robot to be anticipatory instead of being reactive in more handover episodes, two participants expressed preference for more consistency in its reaction time and/or behaviours, three participants suggested more involvement of the robot beyond object exchange, for instance, working on the assembly directly with the person by holding a piece while the person is gluing it to the rest of the birdhouse.

As shown in the questionnaires and interview responses, while having diverse impressions of the robot, overall participants considered it to be an adaptive and helpful collaborator, especially in the assembly and painting stages. Improvement that can benefit the robot's perceived fluency and contribution to the task includes increased participation beyond object handovers and multimodal interaction combining speech and gestures. 

\subsection{Perception of the Collaboration}\label{subsubsec_colab}
\begin{figure}[tb]
  \centering
  \includegraphics[width=0.9\linewidth]{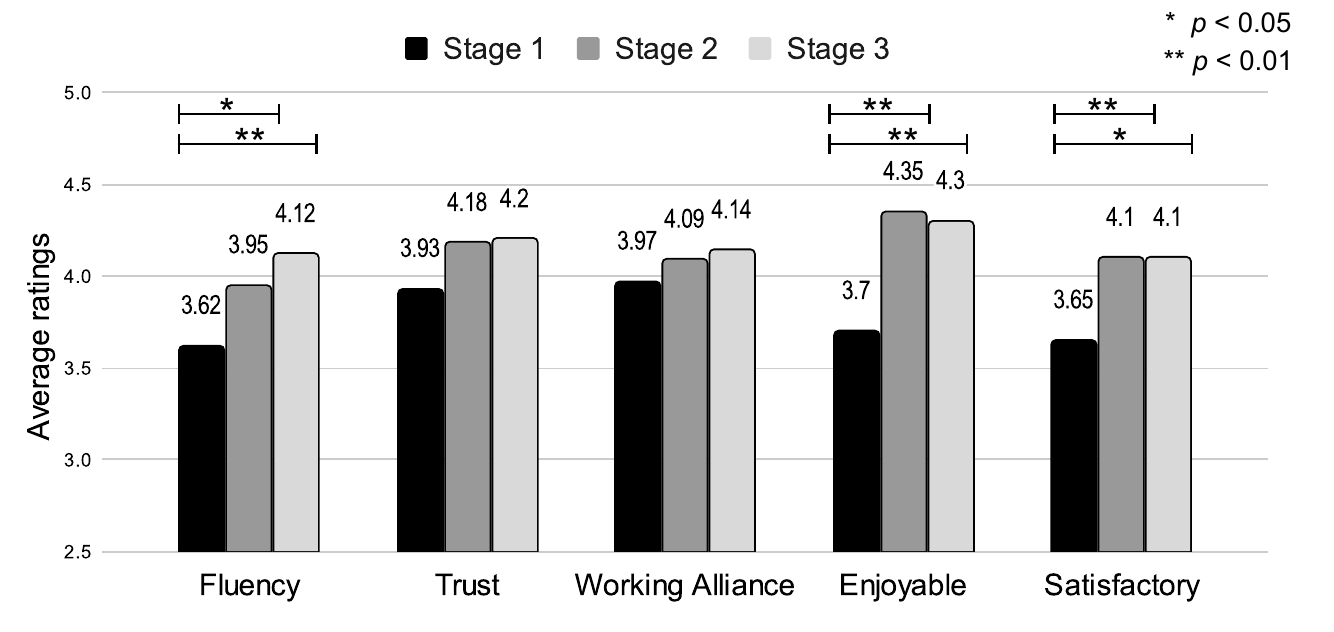}
  % \vspace{-1em}
  \caption{Subjective ratings of HRC: participants perceived the collaboration more positively in Stage 2 \& 3.}
  \Description{Participants have more positive impressions of the HRC in Stage 2 and 3 compared to Stage 1, as measured on the dimensions of fluency, trust towards the robot, working alliance, enjoyable, and satisfactory, with Stage 3 more positive than Stage 2 in terms of fluency.}
  \label{fig:Q_HRC}
  % \vspace{-1em}
\end{figure}

Figure~\ref{fig:Q_HRC} and Table~\ref{tab_qualtrics_robot} showed participants' perception of the HRC. Similar to perception towards the robot, participants have significantly more positive impressions of the HRC in Stage 2 and 3 compared to Stage 1 in terms of Enjoyable, Satisfactory, and Fluency. In the interview, four participants expressed that they found Stage 2 to be the most collaborative out of the three stages where them and the robot worked as an efficient team, while three participants considered Stage 3 to be the most collaborative. 
% Table~\ref{tab_qualtrics_robot} and 
% \footnote{One-tailed t-test showed that the differences are significant ($p<0.05$) for Enjoyable, Satisfactory, and Fluency, but not for Trust and Working Alliance for Stage 1~vs.~2/3.}
% Stage 3 is more positive than Stage 2 in terms of Fluency, Trust, and Working Alliance, although the differences between Stage 2 and 3 are insignificant.
% 4 participants (P1, P2, P6, P20) expressed that they found Stage 2 to be the most collaborative out of the three stages with them and the robot working as an efficient team; 3 (P3, P7, P10) considered Stage 3 to be the most collaborative. Also, 3 participants (P9, P13, P15) found Stage 1 to be the least collaborative while Stage 2 and 3 are somewhat equal and more collaborative than Stage 1. P14 expressed she had no particular preference between the three stages. Further, P2 and P17 considered Stage 2 to be the most enjoyable or fun while P18 found Stage 3 to be more enjoyable.

Participants discussed both positive and negative perceptions of the HRC in interviews. Four participants commented that the experience was ``fun'', for instance, P17 said the wooden pieces in Stage 2 ``arrived like presents''; six expressed that the interaction was ``smooth'', ``efficient'', or ``effective''; five considered the session was ``adaptive'' with the robot improving over time to better address their needs. However, five participants also mentioned that when the robot was waiting for them, especially at the work space, it created a sense of pressure or a desire to speed up their actions. 
% In terms of positive perceptions, 4 participants (P1, P11, P12, P17) commented that the experience was ``fun'', for instance, P17 said the wooden pieces in the assembly stage ``arrived like presents''; 6 participants (P2, P5, P9, P10, P13, P15) expressed that the interaction was ``smooth'', ``efficient'', or ``effective''; 5 participants (P5, P8, P13, P15, P16) considered the session was ``adaptive'' with the robot improving over time to better address their needs. Regarding negative perceptions, 5 participants (P5, P7, P8, P15, P18) mentioned that when the robot was waiting for them, especially at the work space, it created a sense of pressure or a desire to speed up their own actions. 

Participants can also comment in the free text area of the questionnaire after each stage. For Stage 1, four participants commented about their uncertainty in how to signal to the robot to come and to perform object transfer, and their experience gaining better understanding of how the robot will behave over time. For Stage 2, six participants gave comments that highlighted they had a better collaboration experience compared to Stage 1 as the robot felt more involved, efficient, and adaptive. For Stage 3, six participants gave comments discussing their positive impression of feeling relaxed, as well as negative impressions, namely part of the interaction being slow, wanting the objects to be delivered in a different sequence, and the robot handing objects at an undesirable location in some episodes. At the end of the questionnaire, nine participants gave additional comments, expressing both positive impressions of the robot being human-like and adaptive, and negative impressions of the interaction feeling slow or being unsure about the robot's intention at times.
% Participants can also comment in the free text area of the questionnaire after each stage. For Stage 1, 4 participants (P2, P4, P9, P17) commented about their uncertainty in how to signal to the robot to come and to perform object transfer, as well as gaining better understanding of how the robot will behave or react over time. For Stage 2, 6 participants (P1, P2, P4, P6, P17, P18) gave comments that highlighted they had a better collaboration experience compared to Stage 1 as the robot felt more involved, efficient, and adaptive, while P2 also discussed inconsistency in the robot's behaviours as it displayed both anticipatory and reactive handovers. For Stage 3, 6 participants (P1, P2, P3, P4, P9, P17) gave comments discussing their positive impression of feeling relaxed (P3) as well as negative impressions including part of the interaction being slow (P1 and P4), wanting the objects to be delivered in a different sequence (P4 and P17), and the robot handing objects at an undesirable location (P2 and P9). At the end of the questionnaire, 9 participants (P2, P4, P7, P9, P11, P12, P14, P17, P19) gave comments in the free text area, expressing both positive impressions of the robot being human-like and adaptive, and negative impressions of the interaction feeling slow or being unsure about the robot's intention at times.

In the interview, twelve participants noticed the robot changing the OTPs during their sessions, five participants did not notice any location changes, while all handover episodes were at the central location in the sessions with P13, P14, and P19. In addition, three participants noticed the robot changing its arm stretching distance to be closer or further away from the person. Nine participants addressed that the robot was proactive during some episodes, while eleven participants had the impression that the robot was entirely reactive and only initiated handovers when they signalled to it. Two participants also discussed that they were adapting to the robot themselves and became more in tune or familiar with the robot's behaviours as the session progressed.

Our analysis shows that overall participants considered the collaborative experience to be enjoyable, efficient, and adaptive, especially in the assembly and painting stages, despite not noticing all the handover timing and OTP adaptations. 
%The operator's timing and OTP adaptations contributed to this positive and fluent experience. 
% Including more flexibility in the collaboration and having a combination of proactive and reactive collaborations can improve the participants' experience.

\subsection{Functional and Creative Activities}
In the interview, participants discussed how the collaboration with the robot influenced their functional and creative activities during the crafting session. Regarding the functional tasks of assembling the birdhouses, P1 commented that the robot helped avoid cluttered work space. Similarly, P13, P15, and P17 expressed that the robot bringing the wooden pieces one by one guided them to work through an unfamiliar task in a structured manner. Further, P8 referred to her past experience attending furniture design classes, in which people spent long time waiting to use shared equipment, and commented that the robot could be helpful in this scenario by bringing objects to the shared equipment for processing while the person continues with other tasks. 
% P3, P17, and P18 on the other hand commented that as the assembly task is relatively simple (a 7-piece design), having the robot bring all the pieces at the start of assembly would be more efficient.

Regarding the creative activities of designing the appearance of the birdhouse, all participants conceptualised the design during the experiment. Since only the five prime colours were offered, six participants used their phones during the painting stage to search for colour mixing methods (e.g., P2 searched how to create brown) or pictures of specific items that they intended to include in the design (e.g., P9 searched photos of the superb fairywren). Nine participants stated that the order in which the robot delivered the paints directly inspired or influenced their design, while eleven participants commented that as from the instruction sheet they already knew which colours would be provided, their design was not influenced by the order in which the colours arrived. 
% Since only the five prime colours were offered, 6 participants (P2, P9, P11, P17, P18, P20) used their phones during the painting stage to search for colour mixing methods (e.g., P2 searched how to create brown) or pictures of specific items that they intended to include in the design (e.g., P9 searched photos of the superb fairywren). 9 participants (P1, P5, P6, P7, P8, P10, P15, P17, P18) stated that the order in which the robot delivered the paints directly inspired or influenced their design, while 11 participants (P2, P3, P4, P9, P11, P12, P13, P14, P16, P19, P20) commented that as from the instruction sheet they already knew which colours would be provided, their design was not influenced by the order in which the colours arrived. 
% Figure~\ref{fig:teaser} offers an overview of the different designs created by our participants.

Moreover, six participants expressed that they wished the session could have been longer as they were deeply invested in the crafting. Three participants commented that as the human-robot handovers were integrated as part of the crafting session instead of being at the centre of attention, they were able to focus on the creative design process. Specifically, P15 discussed that as he was able to focus on the creative aspect, he recalled the birdhouse's purpose of feeding small birds and changed his design to a more camouflaged look by avoiding bright colours that may attract predators. P10 referred to her past experience interacting with small swarm painting robots, during which the robots made many mistakes and she was concerned that they would break something or hurt someone. However, in this session, she found the robot ``smooth'' and ``helpful'', which allowed her to concentrate on creating and experimenting with different colour mixes. Further, P8 commented from her experiences tutoring children's art classes, where tidying up the supplies at the end of a class is often time consuming. She expressed that the robot helped structure the painting stage to be more organised by delivering and retrieving items in order, which can potentially reduce the workload of the tutors in the children's art class scenario.
% 6 participants (P1, P2, P4, P13, P16, P20) expressed that they wished the session could have been longer as they were deeply invested in the crafting. P2, P10, and P15 commented that as the human-robot handovers were integrated as part of the crafting session instead of being at the centre of attention, they were able to focus on the creative design process. Specifically, P15 discussed that during painting, as he was able to focus on designing the appearance of the birdhouse, he remembered that its goal was for feeding small birds, which led to him changing his design to a more camouflaged look and avoided using bright colours that may attract predators.

As shown in the interviews, participants' functional and creative activities were influenced by the robot's behaviours. The functional collaboration of object handovers not only benefited the person's task efficiency, but also led to design and creative benefits.

\section{Conclusions}\label{sec:dis}
We studied human-robot handovers in a naturalistic HRC context, where a WoZ-controlled mobile robot assisted users in functional and creative tasks. Our analysis of the collected FACT HRC dataset showed that social cues, especially gaze and gestures, are informative for temporal and spatial adaptation in handovers. Further, handovers are shaped by the pre- and post-handover task contexts. Our findings can be generalised to other HRC tasks, such as a cooking assistance robot, to develop socially aware robot behaviours. In the next step, we plan to train an automatic handover model from the collected dataset, which predicts the operator's controls from the social and task context.
% The supervised learning experiments predicting the human operator's controls during WoZ HRC based on our analysis offer initial evidence towards future development of autonomous handover models for more adaptive and socially appropriate HRC.
% Study II and III as future studies (to be included in a separate THRI paper)

\begin{acks}
This work is funded by the Australian Research Council Future Fellowship FT200100761.
\end{acks}

\bibliographystyle{ACM-Reference-Format}
\balance
\bibliography{HRI2023}

% \appendix
% TBA

\end{document}